\documentclass[journal]{IEEEtran}
\usepackage{graphicx}
\usepackage{multirow}
\usepackage{hyperref}
\newcommand{\MYhref}[3][blue]{\href{#2}{\color{#1}{#3}}}%
\usepackage{xcolor,colortbl}
\usepackage[ruled,vlined]{algorithm2e}
\usepackage{amsmath}
\usepackage{amssymb}

\ifCLASSOPTIONcompsoc
    \usepackage[caption=false, font=normalsize, labelfont=sf, textfont=sf]{subfig}
\else
\usepackage[caption=false, font=footnotesize]{subfig}
\fi
\definecolor{Gray}{gray}{0.25}
\definecolor{red}{rgb}{1.00,0.00,0.00}
\definecolor{blue}{rgb}{0.00,0.00,1.00}
\definecolor{green}{rgb}{0.2,0.80,0.0}
\definecolor{yellow}{rgb}{0.5,0.5,0.0}

\newcommand{\cblue}[1] {\textcolor{blue}{#1}}

\usepackage{flushend} 

\ifCLASSINFOpdf
\else
\fi

\begin{document}
%
\title{3D\_DEN: Open-ended 3D Object Recognition using Dynamically Expandable Networks}
%
%
%

\author{Sudhakaran Jain and Hamidreza Kasaei
    \thanks{All authors are with Department of Artificial Intelligence, University of Groningen, PO Box 407, 9700 AK, Groningen, The Netherlands. 
    e-mail: s.j.jain@student.rug.nl, hamidreza.kasaei@rug.nl}
    \thanks{We are grateful to the NVIDIA corporation for supporting our research through the NVIDIA GPU Grant Program.}}



\maketitle

\begin{abstract}
Service robots, in general, have to work independently and adapt to the dynamic changes happening in the environment in real-time. One important aspect in such scenarios is to continually learn to recognize newer object categories when they become available. This combines two main research problems namely continual learning and 3D object recognition. Most of the existing research approaches include the use of deep Convolutional Neural Networks (CNNs) focusing on image datasets. A modified approach might be needed for continually learning 3D object categories. A major concern in using CNNs is the problem of catastrophic forgetting when a model tries to learn a new task. Despite various proposed solutions to mitigate this problem, there still exist some downsides of such solutions, e.g., computational complexity, especially when learning substantial number of tasks. These downsides can pose major problems in robotic scenarios where real-time response plays an essential role. Towards addressing this challenge, we propose a new deep transfer learning approach based on a dynamic architectural method to make robots capable of open-ended learning about new 3D object categories. Furthermore, we make sure that the mentioned downsides are minimized to a great extent. Experimental results showed that the proposed model outperformed state-of-the-art approaches with regards to accuracy and also substantially minimizes computational overhead. Code is available online at: \href{https://github.com/sudhakaranjain/3D_DEN}{\texttt{\small \cblue{https://github.com/sudhakaranjain/3D\_DEN}}}
\end{abstract}


\IEEEpeerreviewmaketitle

\section{Introduction}
\label{introduction}
%
%
%
%

\IEEEPARstart{S}{ervice} robots are generally used in domestic environments where they have to work independently on some given tasks. The structure of such environments are not pre-defined and can be changing frequently. Therefore, robots operating in such dynamic environments should be able to adapt to new changes efficiently overtime. In general, an ideal service robot should intelligently perceive the events occurring in its surroundings and act accordingly. One of the main concerns in such scenarios is learning about new object categories in an open-ended fashion using a few training instances of the object categories. Such a learning process is also called open-ended/continual or lifelong learning~\cite{thrun95k}. It poses a big challenge in the field of Artificial General Intelligence (AGI). Hence, it can be seen as a vital ability required for a robot to perform its day-to-day work. 

\begin{figure*}[!t]
\centering
  \includegraphics[width=1\textwidth, height=0.35\textwidth]{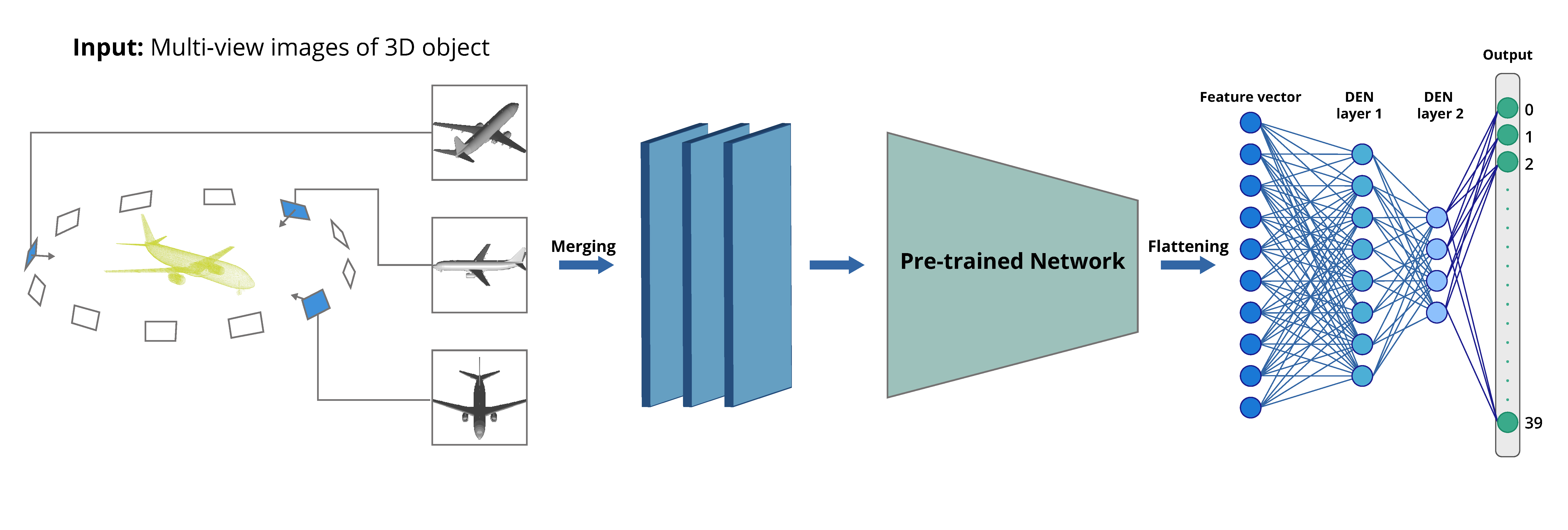}
  \vspace{-13mm}
  \caption{An overview of the proposed \texttt{3D\_DEN} model: Initially, three representative views are chosen from a set of multi-view images for a given 3D object. Then, each of them is converted to a single channel (grey-scale) image and later merged to form a 3-channel image. Now, this image is fed to a pre-trained network, and the extracted features are flattened. Finally, we attach two DEN layers to the model which give the output.}
  \label{fig:model_arch}
\end{figure*}

Even though a lot of attention has been given towards Continual Learning (CL) scenarios, researchers have mainly focused to test their proposed model on 2D image datasets like MNIST~\cite{lwf,regularization,DEN,gem,mnist}, and very little attention has been given to the problem of continual learning of 3D objects. Despite the presence of various papers to improve 3D object classification, open-ended recognition of 3D objects still poses a problem that lacks an efficient solution. Therefore, it has a lot of research scope for improvement~\cite{orthographic}.

In recent times, the research community has been giving much attention to deep Convolutional Neural Networks (CNNs) for traditional object recognition tasks~\cite{review_paper}. Nowadays, it is clear that when the number of object categories are pre-determined (or fixed) and a large number of training instances for each category is available that also resemble the test set, CNNs can give satisfactory results~\cite{orthographic}. Networks can be considered incremental by nature but not open-ended since adding a new output to the network enforces changes in the architecture of the network. In particular, we can incrementally train the network by giving new instances of known categories at each training stage. This makes the model to improve its existing knowledge about known categories with each new training instances. On the contrary, CNNs do not readily support open-ended learning, as this demands new tasks to be learned which may or may not be related to previously known tasks, and thus making the above training approach non-applicable. To be specific, in some open-ended scenarios, the model needs to learn newer categories (here, considered a task) using a few training examples presented over a period of time. This mainly leads to overwriting the network's learned weights every time a new task comes into play. This indeed, causes Catastrophic Forgetting~\cite{catastrophic}, a major limitation of CNNs. The problem can be defined as forgetting the previous tasks while the model is being trained further on newer tasks~\cite{catastrophic}. It usually happens due to overwriting the values of weights learned from previous tasks, upon the start of training on newer tasks. In addition to the problem above, learning to recognize 3D objects adds more complications to the whole issue.

We try to overcome the above-mentioned issues by proposing a novel deep learning dynamic architectural model, and methods to train it that improves open-ended 3D object recognition especially for robotic scenarios. In this paper, we aim to propose a deep transfer learning model with dynamically expandable network (DEN) that is capable of learning a substantially large number of 3D object categories overtime without catastrophically forgetting already known ones. Moreover, we also seek to reduce the computational cost of our model as real-time response is an essential factor for robots. Figure~\ref{fig:model_arch} provides an overview of the proposed approach.

The remainder of this paper is organized in the following way. Section~\ref{section:related} reviews related work of continual learning and 3D object recognition. Next, the detailed methodologies of our proposed model are explained in Section~\ref{section:methodology}. Section~\ref{section:experiment} is about the experimental setup, results and discussion where we explain in-detail about the performance of our model compared to existing state-of-the-art. This is followed by conclusions and future works in the Section~\ref{section:conclusion}.

\section{Related Work}
\label{section:related}

In this study we are looking at the problem of open-ended 3D object recognition, which in itself has two sub-problems, namely continual learning and 3D object recognition. Both of these have a deep history of research in machine learning, computer vision and robotics, resulting in many different approaches. In this section, we review a few recent efforts.

\subsection{Continual Learning Approaches}

Several approaches have been proposed to tackle continual learning problem, involving \emph{regularization techniques} for weight changes, \emph{dynamic architectural networks} and \emph{memory replay}~\cite{review_paper}. \\

\textbf{Regularization Techniques:} This technique tries to tackle the catastrophic forgetting problem by introducing constraint checks on updation of weight parameters of the model~\cite{review_paper}. The use of simple L2-Regularization forbids the model to learn new upcoming tasks efficiently. To overcome this problem, a method called \textit{Elastic Weight Consolidation} (EWC) was proposed in the paper~\cite{regularization} which computes the Fisher information matrix and uses its diagonal to constrain the weight parameters. The paper states that this method identifies which parameters were more important for the previous task and thereby selectively assigns greater constrain on updation of those parameters. This is in contrast to L2-Regularization where we assume an equal constraint for all the parameters. Another research~\cite{si}, came up with a similar strategy called \textit{Synaptic Intelligence} (SI). In this paper, rather than using diagonals of the Fisher matrix to gain information about variance, the authors tried to find the 'importance factor' of each parameter across every previously learned task. This meant that, while updating the parameters the constraint was not just applied to the change of parameters concerning the immediate previous task. Instead, they took into account the weight changes concerning all previously learned tasks. 

X. He et al.~\cite{conceptor_1}, tried to use Conceptors~\cite{conceptor_Jaeger, conceptor_hj} to eradicate catastrophic forgetting. Here, the authors treat conceptors as a memory to learn new tasks, where they keep exploiting only the unexplored part of memory to store the incremental weight parameters of new tasks. \\

\textbf{Architectural Techniques:} This technique tackles catastrophic forgetting by changing the architecture of the model. One such work was the \textit{Progressive Neural Networks} (PNN)~\cite{progressive}, where the authors proposed an idea to increase the size of every layer in the model for each new task. Hence, while performing a task, only the existing nodes and newly added nodes that were present while training that task will be responsible to compute the task's output. But, the main drawback was that the model size kept on increasing with new incoming tasks. Also, as all the tasks used the nodes added only till their training stage, only a sub-network from the whole model was used during inference of each task. This leads to significant wastage of storage space~\cite{DEN}. Extending this work, T. Xiao et al.~\cite{dynArc} proposed a similar model which not only increases its layer sizes but also branches out towards the upper layers. As described by the authors, the model indeed forms a hierarchical structure to accommodate new categories~\cite{dynArc}. Referring to these stated approaches, a new efficient model was recently proposed in the paper~\cite{DEN}. The authors called it `Dynamically Expandable Networks', which combines both regularization and dynamic architectural methods. This paper has proved that by using the three different training methods, better results can be achieved not only concerning accuracy but also in terms of computational complexity; \textit{Selective Re-training}: here, only a sub-network from the whole model is selected and trained on new upcoming tasks; \textit{Dynamic Network Expansion}: the model is expanded dynamically when the loss obtained in the previous method exceeds a specific threshold. Unused newly added nodes are removed after the training process; \textit{Network Splitting}: if the value of an existing node undergoes a drastic change during training, it is duplicated and this new one is used in the further training process. The authors of this paper have also made sure of efficient memory usage without compromising the overall accuracy. This model was tested on several datasets, including 2D image datasets with smaller dimensions like MNIST~\cite{mnist} and its variations. One major disadvantage of using all the above-mentioned approaches is that they all assume each task to be a multi-class classification problem. Due to this, during inference, we have to provide a task ID to select the sub-network from the whole model which may then be used to provide output. Not all continual learning problems will have this scenario, such as our problem statement, where each new task is a new upcoming category to be learned. However, these papers form an excellent basis for our work. \\

\textbf{Memory Replay Techniques:} 
This technique tries to tackle catastrophic forgetting by re-using the old samples belonging to the previously learned tasks. It can further be divided into \emph{rehearsal} and \emph{generative replay}~\cite{Timoth}. According to the paper~\cite{Timoth}, during rehearsal the samples are selected by careful inspection as each set of samples for the previous tasks should thoroughly represent their tasks. However, they can also be chosen randomly. Typically, these samples along with the data of the current task form the core set~\cite{Timoth}. The authors of the paper~\cite{gem} proposed a methodology called 'Gradient Episodic Memory' (GEM), which makes use of this core set not only for training but also to apply a constrain on gradient-update while learning new tasks. To be specific, they try to reduce the angle between the loss gradient vector obtained from samples of previously learned task and the current gradient update~\cite{gem}.
The difference in the generative replay is that we make use of generative models known as GANs~\cite{GAN} to create new samples similar to the original samples of previous tasks. This indeed saves a lot of memory by avoiding the need for previous tasks' data, but adds a small computational cost to train generative models. This cost is mainly due to the usage of two models namely, generator and discriminator~\cite{GAN}. As described in~\cite{Timoth}, the generator has its weights frozen but is used to generate replay samples depicting past experiences. Whereas, discriminator tries to learn from samples of the current task as well as generated replay. When a task is over, the generator model is trained to generate the current task samples, initializing it for the next task. 
Further, the generative replay can be subdivided into `Marginal Replay' and  `Conditional Replay'~\cite{Timoth, gen_replay_types}. According to~\cite{Timoth}, marginal replays make use of conventional GANs, while conditional replays use GANs which accept specific inputs. This means that these models can generate specific output samples as required based on the given input condition. Hence, they can be used to generate samples of a specific previously learned task, if required.

\subsection{Object Recognition Approaches}

There have been a few approaches specific to tackle the problem of open-ended object recognition. One such method is the instance-based approach proposed in the paper~\cite{ergocentric}. This demands less training time and works well when we have fewer samples in training data. According to the algorithm discussed in this paper, the model initially considers each new object as a new class and later clusters them together rather than assigning labels to each sample. Even though the authors have proposed this model for unsupervised continual learning in real-world scenarios, it can also be extended for supervised learning tasks with some modifications.

Another similar instance-based learning model called OrthographicNet was proposed in the paper~\cite{orthographic}. Here, the model learns interactively, based on three functions: Teach, Ask, and Correct. During training, the model tries to generate a global feature vector for every object category, which is scale and rotation invariant. During inference, the extracted features are compared to these global features to recognize the correct category of the input object.

As described in OrthographicNet~\cite{orthographic}, initially, three orthographic projection views of an object are obtained. These views are decided based on the principle component axes obtained from the eigenvalue decomposition of the object's point-cloud. Each view of an object is given as input to a separate pre-trained CNN, which was trained on ImageNet~\cite{imagenet}. This extracts the required features of each view of the object. Then, these features are merged by element-wise max-pooling to form a single representative feature vector for the given object. During training, when an object from a new category arrives, a global representation for that category is created and initialized. This global representation for the category is updated whenever more objects from this category arrive. In other words, each category has its feature representation (in lesser dimension) built up using the instances of that category. Finally, during inference, the feature representation of the input object is compared with the global representations of all categories using a similarity distance measure and is assigned the closest category. This model achieves state-of-the-art accuracy in open-ended 3D object recognition scenarios when tested on \textit{Princeton ModelNet40}~\cite{modelnet} and \textit{Washington RGB-D Object}~\cite{rgb-d} datasets. The main drawback of such an instance-based learning model is that the recognition time increases overtime by learning about new instances. Another concern may be a decrease in performance when training on similar looking categories. This may indeed be a major concern when the number of known categories becomes very high. Moreover, it also occupies memory to store global representations of all categories along with their object views. This problem is mainly due to the usage of a fixed architectural model with a constant number of neurons for feature extraction and representation. Nevertheless, this work serves as one of the major parent papers that inspired our work.

Qi et al.,~\cite{pointnet} proposed a new model named \textit{PointNet} to recognize 3D objects by taking the input of objects in the form of point-cloud. Although this model cannot deal with open-ended learning scenarios, it serves as a base architecture for many other research papers on open-ended learning. This paper describes three key modules: the max-pooling layer also called symmetric function, a local and global information combination structure, and two joint alignment networks~\cite{pointnet}. The Symmetry function makes the model translation invariant and also binds all the information from different points together. The joint alignment network uses an affine transformation matrix on the input before training, so that the model is capable of generalizing certain geometric transformations while classifying. The local features are aggregated using max-pooling to obtain global features. Finally, a non-linear function transforms it to get the corresponding probabilistic output.

Our work differs from the above mentioned papers~\cite{DEN, progressive} in various aspects. First key difference is the type of tasks learned by the model. As stated earlier, our work tries to address the problem where training a new task means to learn and classify a new category with existing known ones, whereas these papers assume every new task to be a multi-class classification that are independent of each other. Secondly, compared to~\cite{DEN} which re-trains all the parameters in the network, we only re-train the newly added nodes during \textit{Dynamic Expansion} to prevent catastrophic forgetting. Hence, we do not need a separate method named \textit{Network Splitting} discussed in~\cite{DEN} to avoid any drastic shift in old parameter values. Lastly, we employ deep transfer learning approach by using a pre-trained network in our architecture for feature extraction instead of training deep convolution layers from scratch.

\section{Methodology}
\label{section:methodology}

Open-ended learning concerns all three issues namely time, computation, and space complexity. Besides, dynamic architectures which we will be using, are known to have very high time/computational cost. However, as our problem at hand is specific to robotics domain, we can assume that the space complexity (storage requirement for data samples) to be less significant. Our main focus is to reduce the computational cost of our dynamic architectural model for faster real-time responses, with little or no compromise in overall accuracy. In other words, we aim to make a model that requires lesser computations during training but gives real-time responses that have acceptable accuracy. 

Similar to OrthographicNet, \texttt{3D\_DEN} takes three different angular view images of each 3D object instance during training. But, rather than having different CNNs for every view image, these view images are converted to grey-scale single-channel images and then combined to form a 3-channel image which becomes our input. Thus, our overall model design is a modified version of OrthographicNet's architecture, that consist of a single pre-trained network (with their weights frozen and top-most layer removed) attached to two consecutive trainable DEN layers (this part was proposed by me) which are Fully-Connected (FC) as shown in Figure~\ref{fig:model_arch}.

By using a pre-trained network and having only one CNN will minimize the training computations to a great extent as this pre-trained network is never trained during our whole open-ended learning process. Whereas, a better overall accuracy will be maintained by DEN layers as they will learn new distinguishing features of upcoming categories during training. Throughout the training procedure, weights connecting these layers are kept sparse using L1-Regularization. However, to achieve an absolute sparseness, we also perform a search at the end of each training process, to find weights having very low absolute value and zero them out.

We introduce modern supervised training methods for training these layers, few of them similar to the ones discussed in the paper~\cite{DEN} with modifications as explained in following subsections. Also, all mathematical equations we formulate are inspired (or modified) from the said paper. The learning procedure consists of a series of tasks. The initial task(\textit{t}=1) is a binary classification of two categories trained according to below Equation~\ref{eq:task_1}: 

\begin{equation}
\underset{\boldsymbol{W}^{t=1}}{\text{minimize}} \,
\mathcal{L}\left(\boldsymbol{W}^{t=1} ; \mathcal{D}_{t=1}\right)+\mu \sum_{l=1}^{L}\left\|\boldsymbol{W}_{l}^{t=1}\right\|_{1}
\label{eq:task_1}
\end{equation}

\noindent where $\mathcal{L}$ is task-specific loss function, $1 \leq l \leq L$ represents the $l_{t h}$ layer in the model. $\boldsymbol{W}_{l}^{t}$ represent the weight matrix of the layer \textit{l} and $\mu$ is the parameter for L1-Regularization. Dataset $\mathcal{D}$ contains image views of $T$ number of object categories, where the set of instances for each task (or category) $t$ is represented by $\mathcal{D}_{t}$. From second task ($t>1$) onwards, each upcoming task represents the addition of a new unknown category to be learned without forgetting already known ones. \\

\noindent \textbf{Data Sampling Procedure:} Ideally, due to the problem of catastrophic forgetting, gradient-based models have to be re-trained again from scratch when newer data becomes available. In our proposed model we try to solve this problem by using the simplest method which is sampled rehearsal of older data along with new data during training each task. This procedure is carried out by sampling the old data for the same number of instances that the new data has. We keep equal sampling probability for all the categories in old data and also maintain a minimum threshold of $\rho$, on the number of instances to be drawn from an old category during sampling. Algorithm~\ref{alg:data_sampling} describes our sampling process. Here, the final data for training task $t$ is represented by $\mathcal{D}_{t}^{'}$ and the $len()$ function returns the size of its input parameter. Although we reduced the training data by using sampling, we still have to store the whole dataset during our open-ended learning process. An ideal solution for generic open-ended learning should eradicate this storage problem. We neglect this problem as we assume our scenario to be purely robotic. \\

\begin{algorithm}[!b]
\SetAlgoLined
\KwIn{Dataset $\mathcal{D}=\left(\mathcal{D}_{1}, \ldots, \mathcal{D}_{T}\right)$, task $t$ $(>1)$}
\KwOut{$\mathcal{D}_{t}^{'}$}
$s=len(\mathcal{D}_{t}) / (t-1)$ \\
 \If{$s<\rho$}{
 $s = \rho$
 }
\For {$i=1,\ldots,t-1$}{
Sample $s$ elements from $\mathcal{D}_{i}$ and add to $\mathcal{D}_{t}^{'}$
}
Add $\mathcal{D}_{t}$ to $\mathcal{D}_{t}^{'}$
\caption{Data Sampling}
\label{alg:data_sampling}
\end{algorithm} 

\begin{figure}[!t]
\centering
  \includegraphics[width=1.0\linewidth]{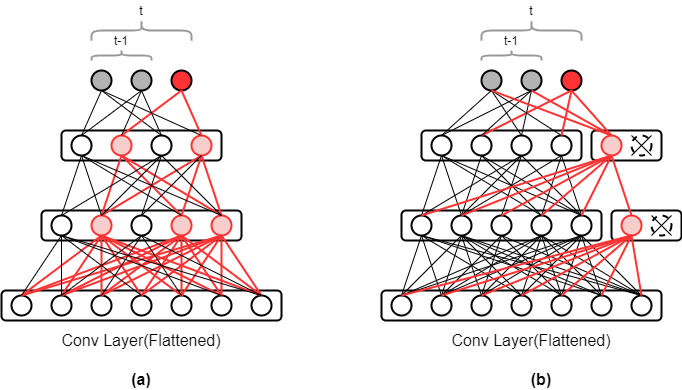}
  \caption{Depiction of training methods used for our model: (a) Selective Retraining: First, the hidden-to-output nodes are selected, then we perform breadth-first search using the selected nodes to identify rest of the nodes that are responsible for new output. Selected sub-network is represented by red color; (b) Dynamic Expansion: New nodes are added to DEN layers and trained, while rest of the nodes are kept constant. In the end, new nodes that were deemed useless are removed. Here, newly added nodes are represented by red color.}
  \label{fig:den}
\end{figure}

\thickmuskip=0.5\thickmuskip
\noindent \textbf{Selective Retraining:} When a new task ($t>1$) arrives, firstly, a new output node $o_{t}$ (for new category) is added and the weights between top-most hidden layer to the output layer (all output nodes) are retrained using the Equation~\ref{eq:select_th}. Then, the nodes in this hidden layer that have non-zero weights to the new output node are selected and added to a list $S$. Further, we perform a layer-wise search among the rest of the sparse DEN-layers (FC) to select all the corresponding nodes which are connected to the previously selected nodes and add it to $S$. These selected nodes are the ones affected by the new task. Hence, we retrain only them separately as a sub-network using the Equation~\ref{eq:train_selected}. Please note that we also make sure there is no drastic change in their weights by imposing a regularization constraint in the loss function. Figure~\ref{fig:den}(a) and Algorithm~\ref{alg:selective_retrain} illustrate the working of this method. 

\begin{equation}
\underset{\boldsymbol{W}_{\boldsymbol{L}, t}}{\text{minimize}} \,
\mathcal{L}\left(\boldsymbol{W}_{L, t}^{t} ; \boldsymbol{W}_{1: L-1}^{t-1}, \mathcal{D}_{t}^{'}\right)+\mu\left\|\boldsymbol{W}_{L, t}^{t}\right\|_{1}
\label{eq:select_th}
\end{equation}

\begin{equation}
\begin{aligned}
\underset{\boldsymbol{W}_{S}}{\operatorname{minimize}} \, \mathcal{L}\left(\boldsymbol{W}_{S}^{t} ; \boldsymbol{W}_{S}^{t-1}, \mathcal{D}_{t}^{'}\right)+\mu\left\|\boldsymbol{W}_{S}^{t}\right\|_{1}\\ +\lambda\left\|\boldsymbol{W}_{S}^{t}-\boldsymbol{W}_{S}^{t-1}\right\|_{1}
\end{aligned}
\label{eq:train_selected}
\end{equation}

\noindent
\noindent where $\boldsymbol{W}_{L, t}^{t}$ represents the weight tensor of nodes at output layer ($L$) which is being trained for task $t$; $\boldsymbol{W}_{1: L-1}^{t-1}$ represents weight tensor of nodes at top-most hidden layer ($L-1$) with values obtained after task $t-1$. $\boldsymbol{W}_{S}^{t}$ represents weight tensor of nodes present in the list $S$, which is being trained for task $t$; likewise, $\boldsymbol{W}_{S}^{t-1}$ represents weight tensor of nodes present in the list $S$ after they were trained for task $t-1$. $\mu$ is the coefficient of L1-Regularization, and $\lambda$ is the coefficient of regularization that governs drastic weight change. \\

\begin{algorithm}[!t]
\SetAlgoLined
\KwIn{$\mathcal{D}_{t}^{'}$, $\textbf{W}^{t-1}$}
\KwOut{$\textbf{W}^{t}$}

Initialize $l \leftarrow L-1, S=\left\{o_{t}\right\}$ \\
Solve Eq. 2 to obtain $\textbf{W}^{t}_{L,t}$ \\
Add neuron $i$ to $S$ if the weight between $i$ and $o_t$ in $\textbf{W}^{t}_{L,t}$ is not zero \\
\For {$l=L-1,\ldots,1$}{
Add neuron $i$ to $S$ if there exists some neuron $j \in S$ such that $\textbf{W}_{l,ij}^{t-1} \neq 0$ \\
}
Solve Eq. 3 to obtain $\textbf{W}_{S}^{t}$
\caption{Selective Retraining 
}
\label{alg:selective_retrain}
\end{algorithm} 

\noindent \textbf{Dynamic Expansion:} When the test accuracy $A_t$ (achieved on unseen test samples) falls below a specific threshold (set as $\tau=0.85$) during selective retraining, then new nodes are added to the model to improve performance. The value of $\tau$ was found empirically through manual trial and error by considering two factors: per-task accuracy and number of new nodes at the end of whole training process. Initially, a constant number of new nodes (say $k$) are added at each layer to train the model for the new object categories. Here, once again the value of $k$ is found through manual trial and error in such a way that an increase in this value does not improve the overall accuracy further. The new nodes have input connections from all nodes from previous layer but their outputs are passed only to new nodes of the next layer. This allows the new nodes to capture new features of the new task category by exploiting information from old nodes without modifying them and thereby preserving the acquired knowledge about older tasks~\cite{progressive}. We only train these new nodes keeping the rest of the network non-trainable, using the below Equation~\ref{eq:dyn_expansion}:

\begin{equation}
\underset{\boldsymbol{W}_{l}^\mathcal{N}}{\operatorname{minimize}} \, \mathcal{L}\left(\boldsymbol{W}_{l}^{\mathcal{N}} ; \boldsymbol{W}_{l}^{t-1}, \mathcal{D}_{t}^{'}\right)+\mu\left\|\boldsymbol{W}_{l}^{\mathcal{N}}\right\|_{1}
\label{eq:dyn_expansion}
\end{equation}

\noindent\noindent where $\boldsymbol{W}_{l}^\mathcal{N}$ represents the weight tensor of newly added nodes at layer \textit{l}; $\boldsymbol{W}_{l}^{t-1}$ represents the weight tensor for nodes at layer $l$ with its values obtained after task $t-1$. Any new node that seems to be useless after training a task is later removed. We assume a node to be useless when the absolute values of all of its output weights are lesser than a specific threshold~$\epsilon$ (set empirically). This makes sure that both the training procedure as well as the size of the model's architecture are optimized with regards to computation cost, at the end of the learning process. Figure~\ref{fig:den}(b) and Algorithm~\ref{alg:dynamic_expansion} illustrate the working of this method.

\begin{algorithm}[!b]
\SetAlgoLined
\KwIn{$\mathcal{D}_{t}^{'}$, $\epsilon$}
\KwOut{$\textbf{W}^{t}$}
Add $k$ neurons $\boldsymbol{\mathit{h}}^\mathcal{N}$ in all layers \\
Solve for Eq. 4 for all layers \\
\For {$l=L-1,\ldots,1$}{
Remove useless neurons in $\boldsymbol{\mathit{h}}^\mathcal{N}_{l}$ whose output weights are less than $\epsilon$
}
\caption{Dynamic Expansion 
}
\label{alg:dynamic_expansion}
\end{algorithm}

\section{Experimental Results}
\label{section:experiment}


Three types of experiments were carried out to evaluate our proposed model. First, we present a systematic open-ended evaluation of the proposed \texttt{3D\_DEN} approach in the context of the object recognition task. Second, we perform an offline evaluation using a similar architecture as \texttt{3D\_DEN}, but with a fixed size of FC-layers. Lastly, we also performed a real-robot demonstration in the context of \texttt{serve\_a\_drink} scenario to show the strength of the proposed approach concerning real-time performance. In the following subsection, for each type of experiment, we first describe the experimental setup and then discuss the obtained results. Open-ended evaluation was performed using both \textit{ModelNet40}~\cite{modelnet} and \textit{RGB-D Object}~\cite{rgb-d} datasets, whereas rest of the experiments were performed only using \textit{ModelNet40}.

\textit{ModelNet40} dataset contains $12,311$ CAD models from $40$ different object categories, which were divided into $9,843$ training samples and $2,468$ testing samples. Image view format of this dataset was obtained from the paper~\cite{mvcnn}. Here, each object sample is represented by a set of $12$ image views that are obtained by setting the camera around the object at different angles on the horizontal plane of the object (angular difference of $30^\circ$ from each other w.r.t object). Each view was given a unique ID (value between $1-12$) to represent its respective camera angle. By manually inspecting a few such samples, we realized which set of camera positions will be sufficient enough to describe the shape of the object. We observed that the image views having camera positions which were approximately $120^{\circ}$ (a circle that lies on a plane forms $360^{\circ}$) apart from each other, best described the shape of the objects. Thereby, we identified and selected the three best representatives (using their unique ID) out of $12$ views for every object, which had the above stated angular difference. Also, note that each view image was converted to a single channel (grey-scale) with dimensions $128\times128$. The starting step illustrated in Figure~\ref{fig:model_arch} might give a better picture of how we obtain the best representative views.

\textit{RGB-D Object} dataset contains $250,000$ views belonging to $300$ common household objects organized into $51$ categories. Here, each object sample is represented by a set of $3$ image views obtained by taking orthographic projections of the object. Each of these view images were converted to single channel (grey-scale) with dimensions $128\times128$. This is how we ultimately formed our final datasets (includes both train and test data). The above pre-processing step was performed to reduce the computational cost of the training process.


\subsection{Open-Ended Training Procedure}

In this round of experiments, we evaluated three different pre-trained networks to find out the best architecture for our \texttt{3D\_DEN} model in terms of accuracy and computational cost. Two of the networks are popular feature extractors namely VGG16~\cite{vgg16} and MobileNet-v2~\cite{mobilenetv2}, both pre-trained on ImageNet~\cite{imagenet}. The third network is a custom model built based on the architecture of MobileNet-v2, and trained on \textit{ModelNet10} dataset~\cite{modelnet} which later serves as a pre-trained network for this experiment. The intention of considering various pre-trained architectures was to find if there is any notable difference in feature extraction when the input for pre-training is not RGB images but rather three single-channel view images. Table~\ref{tab:extractors} gives more insights on these networks' properties. Towards the end, we compare the best \texttt{3D\_DEN} variant to OrthographicNet, and discuss which approach yields better results.

\begin{table}[!t]
\renewcommand{\arraystretch}{1.3}
\caption{Properties of Feature Extractor Networks used in Experiment}
\label{tab:extractors}
\centering
\begin{tabular}{|c | c | c |}
\hline
Model & Feature Length & Depth \\ [0.5ex] 
\hline\hline
VGG16 & $8192$ & 23 \\
\hline
MobileNet-v2 & 1280 & 88 \\
\hline
Custom & 1280 & 88 \\
\hline
\end{tabular}
\end{table}
The training of the model is performed in a supervised manner where the input is a $3$-channel image obtained by combining three grey-scale views of an object (as described in the previous Section), while the label of the object is fed as output. The model makes use of the training methods described in Section~\ref{section:methodology} to learn these input representations and match with the outputs. The optimization happens through a gradient-based stochastic optimizer called Adam~\cite{adam}. We kept the default values for all the hyper-parameters used in this optimization technique.

We conducted this experiment on two datasets namely \textit{ModelNet40} and \textit{RGB-D} in order to concretely validate our proposed model. Since the order of introducing tasks may influence the performance, we performed $10$ trials for each of our \texttt{3D\_DEN} variants on both these datasets. In each trial, the variants were trained from scratch in an open-ended fashion and their overall accuracies were noted down. Every trial using \textit{ModelNet40} consisted of $39$ tasks, whereas trials using \textit{RGB-D} had $50$ tasks. The number of tasks is one less than the total number categories in their respective datasets because the initial task is a binary classification. The order in which these categories appeared for training was random in every trial. Algorithm~\ref{alg:training_procedure} describes this process in a step-wise manner. After the completion of trials, measurements based on different metrics were computed.

\begin{algorithm}[!b]
\SetAlgoLined
\KwIn{Dataset $\mathcal{D}=\left(\mathcal{D}_{1}, \ldots, \mathcal{D}_{T}\right)$}
\KwOut{$\textbf{W}^{T}$}
\For {$t=1,\ldots,T$}{
  \eIf{$t=1$}{
  Train the network weights $\textbf{W}^{1}$ using Eq. 1
  }{
      $\mathcal{D}_{t^{'}}$ = \textit{DataSampling}($\mathcal{D},t$) \\
      $\textbf{W}^{t}$ = \textit{SelectiveRetraining}($\textbf{W}^{t-1}$) \\
      \If{$A_t<\tau$}{
        $\textbf{W}^{t}$ = \textit{DynamicExpansion}($\textbf{W}^{t}$)
      }
  }
}
\caption{Training Procedure (modified from~\cite{DEN})}
\label{alg:training_procedure}
\end{algorithm}

\begin{figure*}[!t]
\centering
  \includegraphics[width=0.95\textwidth]{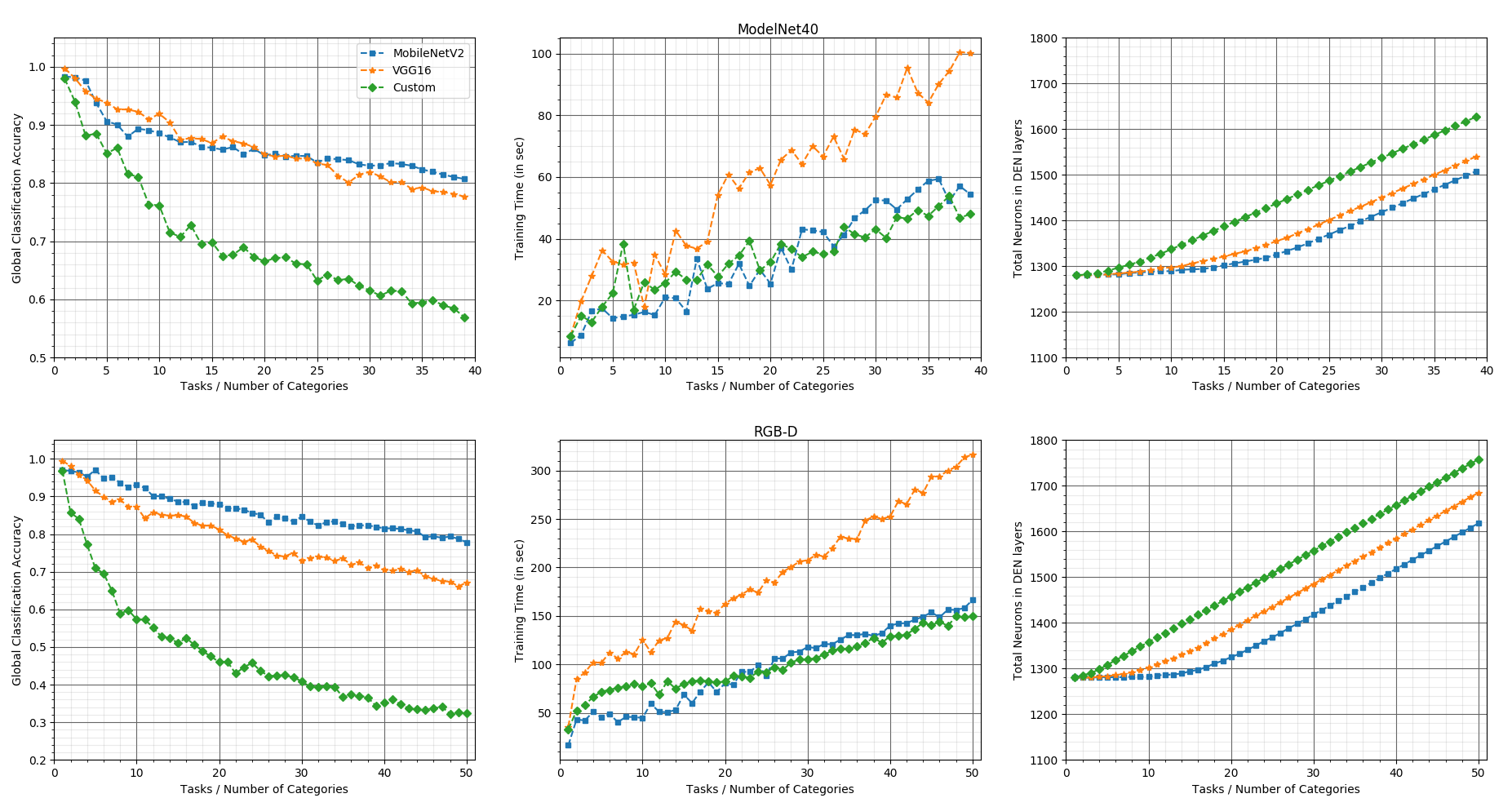}
  \caption{Summary of open-ended evaluation: (\textit{left}) shows the timeline of accuracies for all three models on both datasets. We can notice \texttt{3D\_DEN\_MobileNet} outperforms the rest on both datasets; (\textit{middle}) shows training time (approx) needed by our variants while learning new tasks. Note the difference between \texttt{3D\_DEN\_VGG16} and rest of them, which is due to the size of the pre-trained network used; (\textit{right}) shows the increment in number of neurons in DEN layers for all variants on both datasets. With careful inspection, we can notice that the steady increase starts only after a specific threshold for every \texttt{3D\_DEN} variant.}
  \label{fig:all_graph}
\end{figure*}

It should be noted that the classical form of evaluation which considers accuracy as the main metric cannot be used for open-ended performance evaluation. This is because, in open-ended evaluation, training happens continually and therefore the final accuracy obtained in the end cannot be the only criterion that defines the efficiency of the training methods. Hence, we consider three main metrics introduced by the papers~\cite{metrics, orthographic, kasaei_coping} to compare and discuss the performance of our \texttt{3D\_DEN} variants. These metrics include \textit{(i)} Global Classification Accuracy (GCA), which computes the average of final accuracies for all our trials; \textit{(ii)} Average Protocol Accuracy (APA), which describes the average accuracy over all tasks and all trials; \textit{(iii)} Average number of Learned Categories (ALC) during each trial. In OrthographicNet, a trial is stopped as soon as the test accuracy falls below a given threshold (set as 66.7\%), and the agent can not recover its accuracy after 100 iterations. Therefore, each trial may learn a different number of tasks, and thereby ALC value is computed. But, in our approach we do not have such a stopping criterion, rather we continue to train the model till all categories (tasks) are learned during each trial. Thus, the ALC values for our \texttt{3D\_DEN} variants is $40$ and $51$ on \textit{ModelNet40} and \textit{RGB-D} respectively, which intuitively means that each trial involved open-ended training of all categories from both datasets. This is done to compare the full capability of each \texttt{3D\_DEN} variant in learning new tasks and decide which performs the best. This ALC criterion was used in our evaluation only to compare how our \texttt{3D\_DEN} variants perform with respect to OrthographicNet. Additionally, to evaluate the computational cost of training, we consider the total number of parameters in each of our variants. \\

\begin{table*}[!h]
\renewcommand{\arraystretch}{1.2}
\caption{Results of Open-Ended Evaluation on ModelNet40 and Washington RGB-D object datasets.}
\label{tab:results}
\centering
\resizebox{0.8\textwidth}{!}{
\begin{tabular}{|c | c | c | c | c | c | c | c |}
\hline
    \multirow{2}{*}{Model} &
      \multicolumn{3}{c|}{\textit{ModelNet40}} &
      \multicolumn{3}{c|}{\textit{RGB-D}} &
    \multirow{2}{*}{\#Parameters} \\
\cline{2-7}
 & GCA(\%) & APA(\%) & ALC & GCA(\%) & APA(\%) & ALC & \\ [0.5ex] 
\hline\hline
OrthographicNet~\cite{orthographic} & 66.5 & 74.7 & 38 & 77.0 & 80.0 & 51 & 10.59 M \\ 
\hline
\texttt{3D\_DEN} (ours-\textit{VGG16}) & 77.6 & 83.9 & 40 & 67.2 & 77.2 & 51 & 138.35 M \\
\hline
\texttt{3D\_DEN} (ours-\textit{MobileNet}) & \textbf{80.7} & \textbf{84.2} & 40 & \textbf{77.8} & \textbf{84.7} & 51 & \textbf{3.53 M} \\
\hline
\texttt{3D\_DEN} (ours-\textit{Custom}) & 56.9 & 68.2 & 40 & 32.3 & 46.8 & 51 & 3.53 M \\
\hline
\end{tabular}
}
\end{table*} 

\textbf{Results:} During our extensive set of open-ended training with different settings in the experiment, we compared the performances of our \texttt{3D\_DEN} variants with OrthographicNet (current state-of-the-art) using two datasets. Table~\ref{tab:results} shows the results. We can see that our \texttt{3D\_DEN} variant with Mobilenetv2 as its pre-trained network performed the best in terms of both the accuracies namely GCA and APA, making it the new state-of-the-art model for open-ended evaluation on these datasets. Although, the difference in terms of overall performance compared to OrthographicNet is more distinctive for the \textit{ModelNet40} dataset. Moreover, OrthographicNet performed so poorly on this dataset that it had to be stopped after \textit{task-38}. It should also be noted that our best variant \texttt{3D\_DEN\_MobileNet} showed better metric values on the test data of both datasets while using only one-third of the parameters of OrthographicNet. This indeed says that converting of view images to single channels and combining them drastically reduced the computational time in comparison with OrthographicNet.

Figure~\ref{fig:all_graph} shows the timeline graph of GCA (\textit{left}), training time (\textit{middle}) and increase in number of neurons in DEN layers (\textit{right}) while learning newer tasks evaluated for all \texttt{3D\_DEN} variants on both datasets. From the plots on the \textit{left}, we can see that as the newer tasks are being learned, GCA for all the variants started to decrease gradually. This was an expected phenomenon because when newer knowledge is being gained, the chances of remembering all learned tasks reduces. To put it technically, as the learning progresses, the weight matrices of the model gradually adapt (or overwrite) themselves to learn newer categories thereby forget to recognize features learned from previous categories. Secondly, we do observe some overlapping between the curves of \texttt{3D\_DEN\_MobileNet} and \texttt{3D\_DEN\_VGG16} for some tasks. Also, by referring to APA values from Table~\ref{tab:results}, both of them performed somewhat similar but differed in terms of GCA values when trained on \textit{ModelNet40}. However, our \texttt{3D\_DEN\_Custom} variant performed the poorest of them all on both datasets. This indeed proves that such custom-trained variants cannot serve as a good feature extractor for open-ended learning scenarios of 3D objects.

The calculated training time plotted on the \textit{middle} considers the training time of DEN layers (i.e. trainable parameters) excluding feature extraction time from pre-trained networks. Please note that this time calculation also depends upon other background processes running on the system, thus making this plot only an approximation to get an overview of the computational complexity of the models. Having said that, we do observe the training time of \texttt{3D\_DEN\_VGG16} to be much higher than the other two variants which can indeed be explained by the size of extracted feature vectors used by them. The plots on the \textit{right} shows a gradual increase in the size of DEN layers with newer tasks. With precise examination, we also observe that \texttt{3D\_DEN\_Custom} starts to add newer neurons in the DEN layer much earlier than other models, which in-turn depicts that it was unable to learn new tasks by training with existing neurons. To put it in technical terms, existing neurons proved insufficient to gain newer insights from the extracted features while learning almost every new task. Hence, this variant had to use the second approach that is Dynamic Expansion at a very early stage.

Another major observation is the slope of the plots on the \textit{left} for all the variants. In case of \texttt{3D\_DEN\_MobileNet} and \texttt{3D\_DEN\_VGG16} their slopes looks a bit steep towards the start but later this steepness decreases with upcoming tasks. With further inspection, we can see that these variants have a threshold value on the x-axis after which the steepness decreases more significantly. To interpret the reason behind this phenomenon, lets consider \texttt{3D\_DEN\_MobileNet} for which this threshold is around \textit{task-15} in x-axis for both datasets. By referring to plots on the \textit{right} which describes the increment of neurons during training, \textit{task-15} seems to be the point after which this variant start to add neurons to its DEN layers. This indeed conveys that our second approach which is Dynamic Expansion, yields better results than selective retraining. Even though both seem to suppress catastrophic forgetting, the latter seems to do the job more efficiently. In case of \texttt{3D\_DEN\_Custom}, its threshold is not related to start of Dynamic Expansion approach as this variant still fails miserably at the initial tasks even after adding new neurons. However, it can be incurred from its threshold value that after adding significant number of neurons (or after several Dynamic Expansion steps) the steepness of its slope decreases marginally.

Even though, our best variant performed well in this experiment, it also gave insights on some complications guiding us to work on them in future. This is regarding the dynamic increase in DEN layers. From the plots in the Figure~\ref{fig:all_graph} (\textit{right}), we can see that after a particular task number, the number of neurons kept increasing at a constant rate. This shows that after adding new neurons, useless ones are not being removed at the end of every task. This can open an avenue to optimize the model further. Nevertheless, when we closely inspected the increment of DEN layer sizes in each trial of the experiment, we did observe some rare occasions in which our mechanism eliminated the useless neurons. The reason behind this drawback is that the Dynamic Expansion approach fails to achieve significant sparsity in the weight matrices of newly added neurons.

\subsection{Offline Evaluation using Grid Search}

Through this evaluation procedure, we try to compare an offline trained model with our best performing \texttt{3D\_DEN} variant (\texttt{3D\_DEN\_MobileNet}) which was trained in an open-ended fashion. This experiment was performed using \textit{ModelNet40} dataset. Precisely, we try to find an approximate number of neurons in FC-layers and the appropriate optimizer required for a model that can achieve the best accuracy when trained in an offline fashion on this dataset. We will then observe how the DEN layers of \texttt{3D\_DEN\_MobileNet} differ from the optimal size of FC-layers in this model obtained and also seek to find if there are any differences in optimizer used.

\begin{figure}[!b]
\centering
  \includegraphics[width=1.0\linewidth]{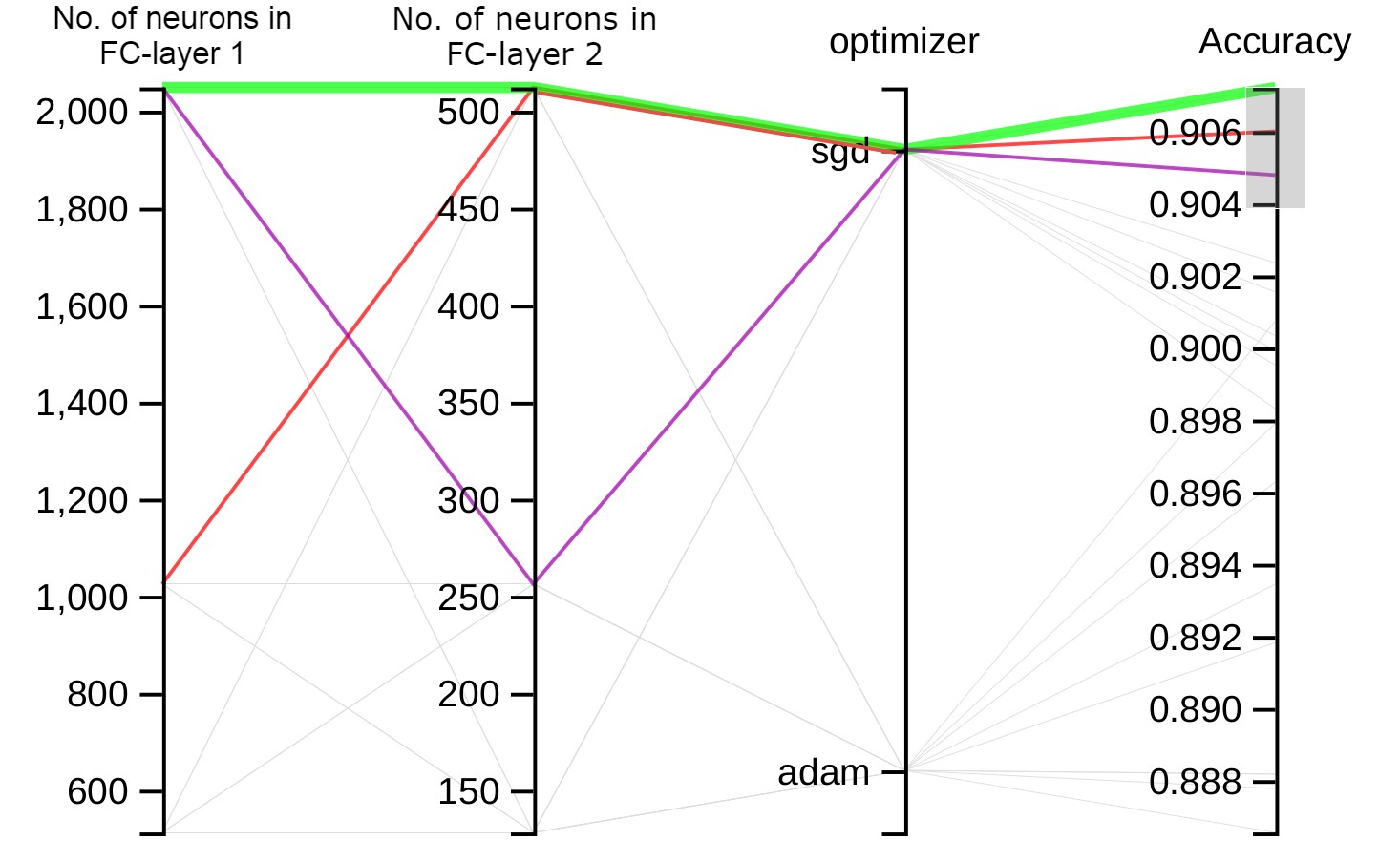}
  \caption{Results of offline evaluation using grid search is shown above. The best parametric configuration is represented by green-colored connection. Whereas, the red and purple-colored connections show the parametric configurations which achieved second best and third best results, respectively.}
  \label{fig:offline_results}
\end{figure}

In this round of experiments, the model training is governed by the Grid Search algorithm. Initially, the parameter range for FC-layer size and set of optimizers to be used (refer Table~\ref{tab:range}) are fed to grid-search. Then, we perform a series of training, each time using the whole dataset and using a new parametric configuration (obtained from grid-search) for the model. In simple terms, each training procedure involves building a model according to a configuration obtained from the grid-search algorithm and training it from scratch on the whole dataset in an offline fashion (on all 40 categories at once). The idea is to inspect which model configuration performs the best as each training procedure has a different combination of FC-layer sizes and optimizer. Table~\ref{tab:range} shows the best configuration results obtained, along with the input parameters of grid-search. The figure~\ref{fig:offline_results} shows all these configurations with their respective accuracies obtained. The top three configurations based on accuracy are shown in different colors (green, red, and purple) while the rest are represented in light-grey. The green-colored connection shows the best configuration. \\

\textbf{Results:} Table~\ref{tab:range} and Figure~\ref{fig:offline_results} shows the results of the offline evaluation, where we can see that the best size of FC-layers were $2048$ and $512$ neurons (i.e., a total of 2560 neurons), yielding the highest accuracy of $90.72\%$. Now, by comparing these results to the performance of \texttt{3D\_DEN\_MobileNet} on \textit{ModelNet40} (referring to Figure~\ref{fig:all_graph}-\textit{right}), we observe that our best variant used only around $1500$ neurons in total for both DEN layers which indeed seems reasonable when we correlate them with the respective accuracies they achieved. Also, to our surprise, we observed that offline evaluation gave the best accuracy using SGD optimizer, whereas open-ended evaluation variants always gave the best results using Adam optimizer.

\begin{table}[!t]
\renewcommand{\arraystretch}{1.3}
\caption{Grid Search Parameters for Offline Evaluation}
\label{tab:range}
\centering
\begin{tabular}{|c|c|c|c|}
\hline
Grid Parameters & Range & Best Parameters & Best Accuracy(\%) \\  
\hline\hline
FC-layer:1 & 512--2048 & 2048 & \multirow{3}*{90.72} \\ 
\cline{1-3}
FC-layer:2 & 128--512 & 512 & \\
\cline{1-3}
Optimizer & SGD, Adam & SGD & \\
\hline
\end{tabular}
\end{table} 

\subsection{Evaluation on Real-Robot}

\begin{figure}[!t]
\centering
  \includegraphics[width=0.8\linewidth, trim= 0cm 0cm 0cm 1.5cm,clip=true]{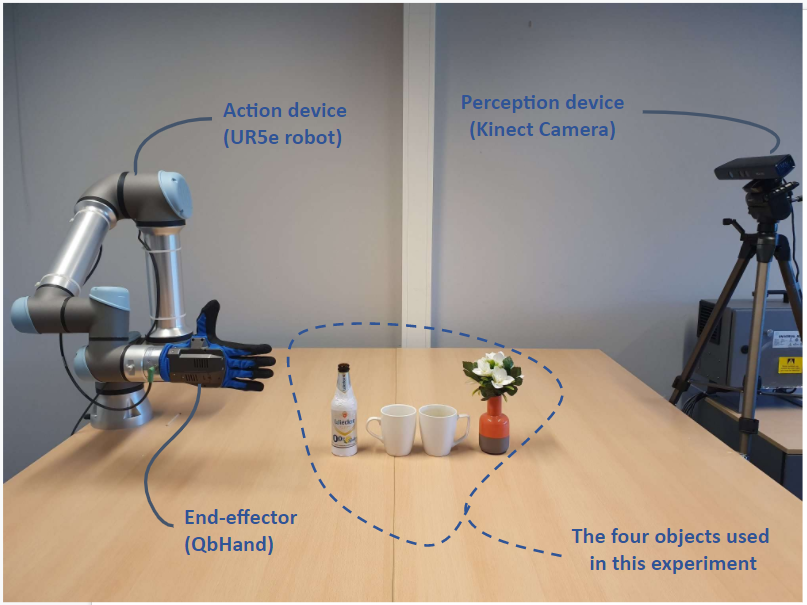}
  \caption{Our experimental setup for real-robot experiment consists of a Kinect sensor to perceive the environment, and a UR5e robot to act upon the environment. Throughout the \texttt{serve\_a\_drink} experiment, we use four instances of three object categories including \textit{bottle}, \textit{cup}, and \textit{plant}. It should be noted that similar instances to the selected objects exist in the \textit{ModelNet40} dataset. }
  \label{fig:robotic_setup}
\end{figure}

A real-robot experiment was carried out to check the performance of our proposed model in real-time. Precisely, this experiment aimed to see if the robot can recognize a set of table-top objects using our best \texttt{3D\_DEN} variant trained on \textit{ModelNet40} to accomplish a given task. In particular, the task here was to pour a drink from a bottle into cups present on the table. The experimental setup is depicted in Figure~\ref{fig:robotic_setup}. It mainly consists of a Kinect sensor and UR5e robotic-arm that perceives the environment and acts accordingly. There are four instances of three object categories on the table: two cups, a bottle, and a vase object with flowers. This is a suitable set of objects for this test since similar instances to these selected objects exist in the \textit{ModelNet40} dataset. Towards this goal, we integrated our trained model into a cognitive robotic system presented in~\cite{kasaei2019interactive} as a ROS service~\cite{ros}. \\

\begin{figure*}[!h]
    \centering
  \subfloat[\label{a}]{%
       \includegraphics[width=0.33\linewidth,trim= 2.5cm 0cm 0cm 0cm,clip=true]{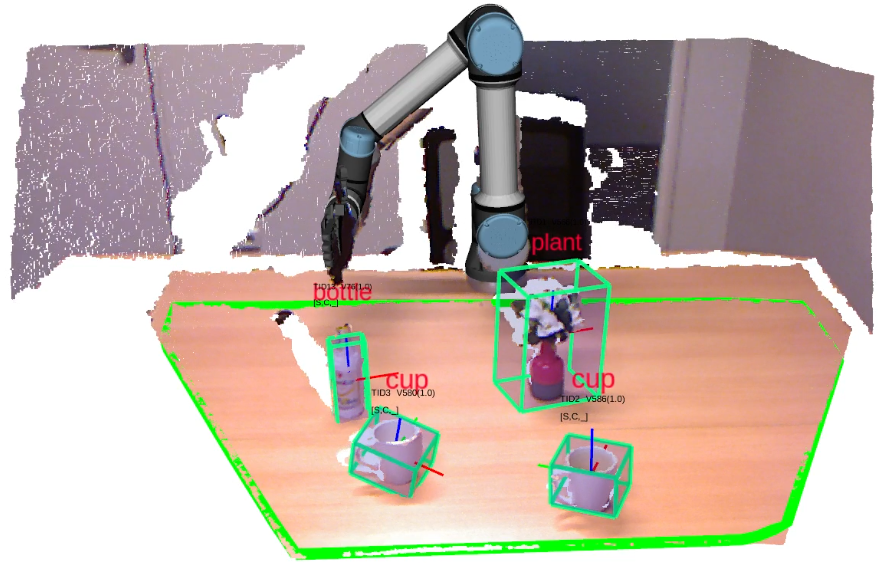}}
    \hfill
  \subfloat[\label{b}]{%
        \includegraphics[width=0.33\linewidth,trim= 2.5cm 0cm 0cm 0cm,clip=true]{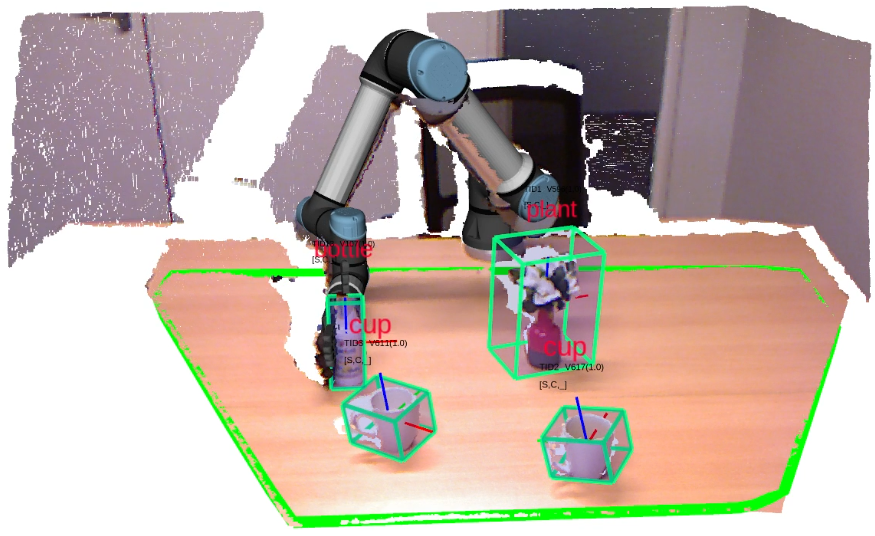}}
    \hfill
  \subfloat[\label{c}]{%
        \includegraphics[width=0.33\linewidth, trim= 2.5cm 0cm 0cm 0cm,clip=true]{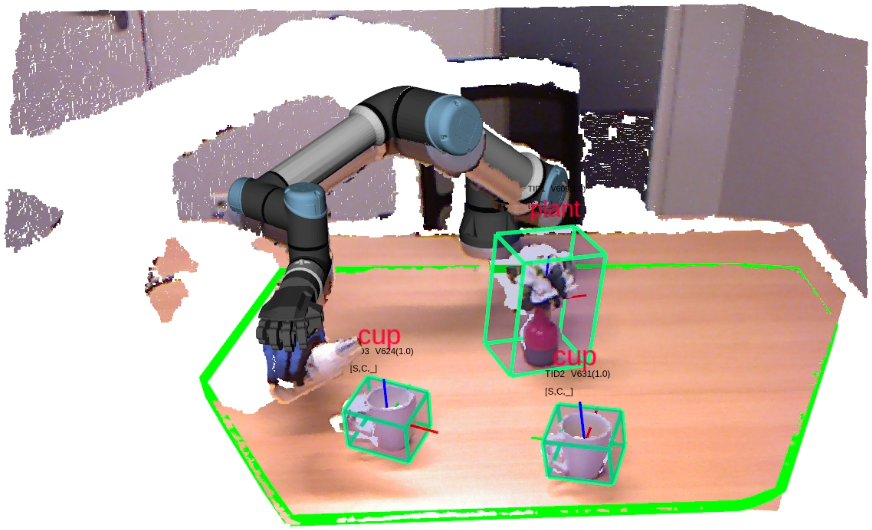}}
  \caption{A sequence of snapshots from the real-robot evaluation: (a) This snapshot shows our model \texttt{3D\_DEN} being used to recognize the objects present on the table; (b) The robot grasps the bottle after recognizing it; (c) The robot manipulates the bottle on top of the cup and tries to pour a drink from the bottle into the cup. In these images, the pose of objects is shown by green 3D bounding boxes, and the recognition results are displayed on top of the objects in red. }
  \label{fig:robot_demo} 
\end{figure*}

\begin{figure}[!t]
\vspace{-5mm}
    \centering
  \subfloat[\label{a}]{%
       \includegraphics[width=0.33\linewidth,trim= 1cm 1cm 1cm 2cm,clip=true]{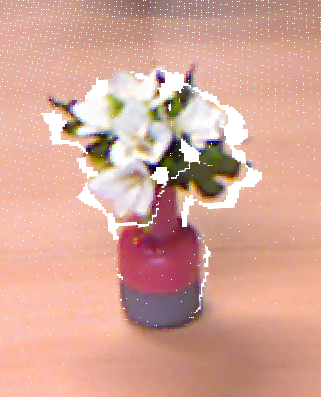}}
    \hfill
  \subfloat[\label{b}]{%
        \includegraphics[width=0.33\linewidth, totalheight=0.14\textheight]{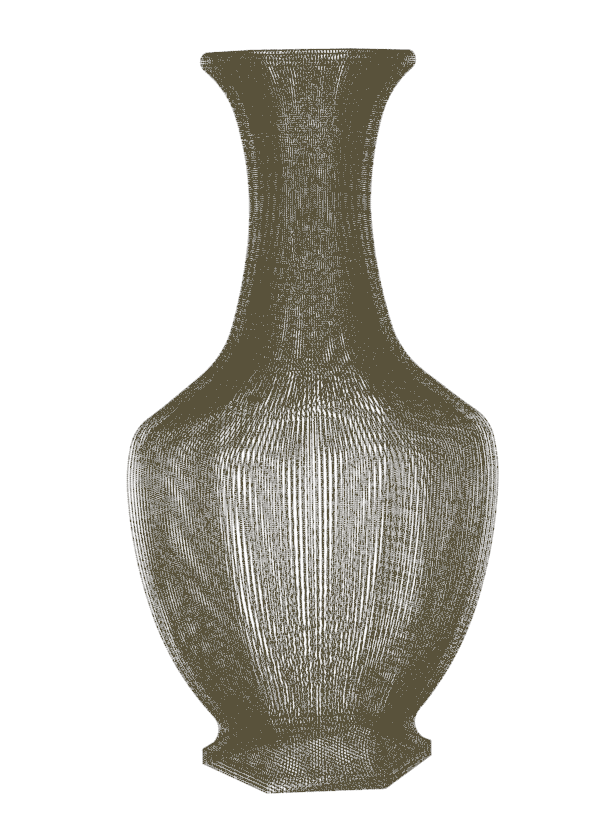}}
    \hfill
  \subfloat[\label{c}]{%
        \includegraphics[width=0.33\linewidth]{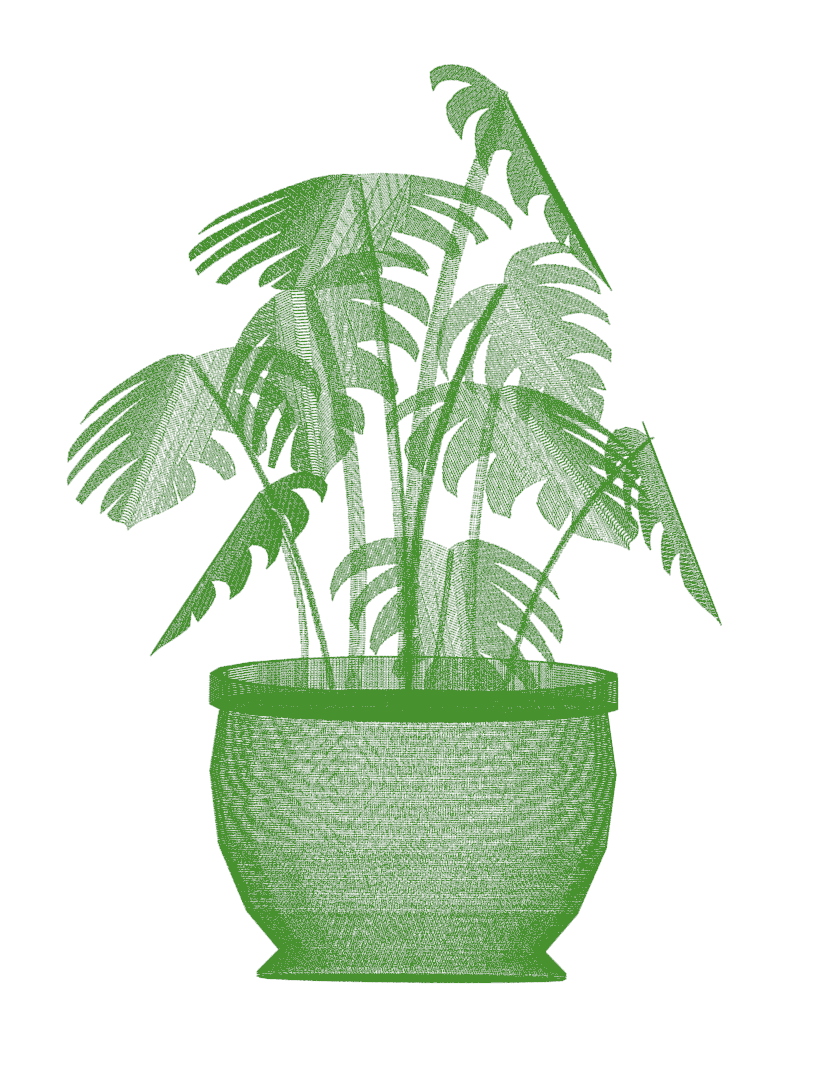}}
  \caption{Comparison between a real-world object with the instances used for training: (a) shows the partial point cloud of the vase object captured by the Kinect sensor in our experiment; (b) and (c) show a vase and plant instance from the \textit{ModelNet40} dataset, on which our model was trained.}
  \label{fig:vase} 
\end{figure}

\textbf{Results:} To accomplish the \texttt{serve\_a\_drink} task, the robot should first detect all the table-top objects, recognize their respective labels, and then decide which affordance action should be performed on objects based on their pose. Afterward, it has to grasp the bottle object and transport it on top of each active cup and serve the drink. The robot should finally return to the initial pose. Towards this goal, the robot first segments the point cloud of each object from the scene and then computes three single-channel orthographic views for each of the objects. Then, the obtained orthographic views of each object are sent to the \texttt{3D\_DEN\_MobileNet} as a service request, and the recognition results are received as the service response (output) in 30Hz. Figure~\ref{fig:robot_demo} shows a sequence of snapshots of the outputs of object recognition and manipulation while the robot performing serve\_a\_drink task\footnote{A video demonstration is available at \MYhref{https://youtu.be/tf4trRMyQ0Y}{https://youtu.be/tf4trRMyQ0Y}}. 
While experimenting with the robot, we observed that the robot was able to precisely detect the pose and recognize the label of all objects most of the time. We also observed that the inferred label for the vase object kept changing between `\textit{vase}' and `\textit{plant}'. This classification error indeed makes sense, because both vase and plant categories from \textit{ModelNet40} dataset (on which \texttt{3D\_DEN\_MobileNet} was trained) look very similar to the vase real-world object. Figure~\ref{fig:vase} shows the comparison between the vase object used during our experiment with vase and plant instances from \textit{ModelNet40}. However, given the good performance of our object recognition model, the robot was able to recognize and grasp the bottle and pour a drink from it into the cups present on the table. \\

\section{Conclusion and Future Scope}
\label{section:conclusion}

In this paper, we proposed a deep learning based approach named \texttt{3D\_DEN} that makes use of dynamic architectural design to learn 3D objects in open-ended fashion. This approach not only achieves better results in terms of accuracy, but also proves to be very efficient in reducing computational cost, which is indeed considered as a major concern in all robotics scenarios that require real-time responses. While this model can be seen as new state-of-the-art benchmark, it should also be noted that this model paves a new path(by using dynamic architecture) for solving the problem of continual learning using 3-dimensional data in robotics domains. It can indeed also be considered in other domains that use 3D data, but of course, with some improvements/changes to adapt to those domains.

In continuation of this work, we would like to consider the possibility of improving the control on dynamic increase in DEN layers. We seek to solve this unfavourable aspect by coming up with a better optimization technique to reduce the usage of neurons. Finding appropriate values for other hyper-parameters (and certain thresholds) is one more future plan. Particularly, we seek to build a strategy which will decide these values using an additional learning-based approach. Another goal is to integrate the ability to extract color features in our model.

\ifCLASSOPTIONcaptionsoff
  \newpage
\fi


%


\bibliographystyle{IEEEtran}
\bibliography{my_references}

%








\end{document}